# Quantifying Phonosemantic Iconicity Distributionally in 6 Languages

**George Flint**
Cognitive Science, UC Berkeley
georgeflint@berkeley.edu

**Kaustubh Kislay**
Data Science, UW Madison
kislay@wisc.edu

## Abstract

Language is, as commonly theorized, largely arbitrary. Yet, systematic relationships between phonetics and semantics have been observed in many specific cases. To what degree could those systematic relationships manifest themselves in large scale, quantitative investigations–both in previously identified and unidentified phenomena? This work undertakes a distributional approach to quantifying phonosemantic iconicity at scale across 6 diverse languages (English, Spanish, Hindi, Finnish, Turkish, and Tamil). In each language, we analyze the alignment of morphemes' phonetic and semantic similarity spaces with a suite of statistical measures, and discover an array of interpretable phonosemantic alignments not previously identified in the literature, along with crosslinguistic patterns. We also analyze 5 previously hypothesized phonosemantic alignments, finding support for some such alignments and mixed results for others.

**Code:** https://github.com/roccoflint/quantifying-iconicity

## 1 Introduction

The relationship between phonetics and semantics in language has been traditionally characterized as predominantly arbitrary (de Saussure, 1983; Hockett, 1960). Concurrently, observations of systematic relationships–phonosemantic iconicity–have counterbalanced these notions. Vainio (2021) found that people systematically match nonsense words to hand actions based on word phonetics, where low sonority words are matched with precision manipulations, and high sonority words are matched with gestures of greater magnitude, such as opening a jar or hammering. Likewise, Ćwiek et al. (2021) found that, crosslinguistically, people systematically match "Kiki" to sharper, spiker shapes and "Bouba" to rounder shapes. Chen et al. (2016) found that US and Taiwanese participants match "Kiki" to spiker radial patterns and "Bouba" to rounder radial patterns. In a descriptive analysis on phonosemantic relationships in the English vocabulary, Bolinger (1950) found networks of words connected through "rime and assonance" (phonesthemes), identifying associations such as /gl-/ relating to visual phenomena or /fl-/ relating to movement phenomena.

Computational rediscoveries of such identified phenomena also exist in the literature. Blasi et al. (2016) discovered a slew of associations between preconfigured sets of phonemes and concepts across thousands of geographically diverse languages. For example, the concept *small* were found to be associated with phonemes such as /i/ or /ʧ/; *round* was associated with /r/. Interestingly, some negative results contradict hypotheses elsewhere in the literature. For example, /o/ was not associated with *big*, as a magnitude–sonority scale alignment might suggest. Liu et al. (2018) develop a feature selection approach using sparse regularization for phonestheme rediscovery, finding that 13 of the 15 model-predicted phonesthemes were previously identified in the literature. Previous computational work has also sought to measure phonosemantic statistical dependency effects at language-scale. de Varda and Strapparava (2022) develop a Long Short-Term Memory (LSTM)-based model successfully learned to predict words' semantic embeddings given phonetic structure after crosslinguistic training, even achieving generalization to an out-of-distribution language.

This work also presents a quantitative approach, at language-scale, measuring global phonosemantic statistical dependencies and investigating both previously hypothesized and unidentified phonosemantic phenomena using distributional methods. We investigate 6 diverse languages (English, Spanish, Hindi, Finnish, Turkish, and Tamil). We conduct a suite of statistical analyses across large sets of morphemes in these languages, and inter-pret identified dimensions of phonosemantic alignment within them. We also investigate 5 proposed phonosemantic alignments: magnitude-sonority (Vainio, 2021), angularity-obstruency (of the famous “Kiki”–“Bouba” effect; Ćwiek et al., 2021), fluidity-continuity (Bolinger, 1950), brightness-vowel frontness (Nuckolls, 1999), and agility-phonological lightness (Berlin, 1995). We find phonosemantic relationships at scale, and find both supportive and contradictory results for proposed phonosemantic alignments. We also discover an array of interpretable phonosemantic alignments and a set of possible interpretations for them.

## 2 Methodology

### 2.1 Language selection

We investigate 6 moderate- to high-resource languages of some typological diversity—3 within the Indo-Eurpoean family and 3 outside of it: English (Indo-European Germanic), Spanish (Indo-European Romance), Hindi (Indo-European Indo-Aryan), Finnish (Uralic), Turkish (Turkic), and Tamil (Dravidian). Typological diversity was maximized under constraints of familiarity of the authors with the languages (for verification purposes).

### 2.2 Preprocessing

#### 2.2.1 Word selection

For each language, we gather the top 5000 words by frequency using the Wordfreq module (Speer, 2022). Wordfreq calculates frequencies from 8 compiled domains of text, including Wikipedia, Subtitles, News, Books, Web text, Twitter, Reddit, and Miscellaneous sources.

#### 2.2.2 Morphological segmentation

For global analyses, comparing phonetic and semantic similarities of items at the word level presents a methodological confound: when two words share a morphological constituent, similarities will align at least in part by mere transitivity. If the items were decomposed into morphemes, their similarity alignments (or lack thereof) could measure true phonosemantic iconicity devoid of the confound. For example, similarities between the words ‘connection’ and ‘construction’ are confounded by shared morphemes ‘con’ and ‘ion.’ Thus, words must undergo both derivational and inflectional morphological segmentation.[1]

This requirement presents challenges to existing tools. Supervised approaches such as Chipmunk (Cotterell et al., 2015) or a character-based neural network approach (as in Pranjić et al., 2024) are not viable due to a lack of labeled data for our selected languages and task. Morfessor (Virpioja et al., 2013)–an unsupervised method–failed to segment derivational morphology. (For example ‘connection’ might be decomposed into ‘connect’ but no further.)

One emerging class of tools for this problem is the use of large language models (LLMs), whose inherent flexibility and instruction-following capability offer an alternative to algorithmic approaches which can struggle with irregularity or task specificity. For example, Pranjić et al. (2024) develop an LLM-based approach to morphological segmentation. However, their approach requires labels and extensive training. One approach which offers flexibility and does not require extensive sets of labels is few-shot prompting (Brown et al., 2020). We opt for this approach, using 10-shot prompts to OpenAI’s GPT-4.1 model, which specifically excels at instruction following[2]. To further bolster instruction following, we use structured outputs in the API[3].

To reduce task complexity, selected words are first lemmatized using the Stanza NLP module (Qi et al., 2020) to truncate some inflectional morphology. Before beginning LLM segmentations, we retrieve IPA transcriptions of these lemmas using the Epitran module (Mortensen et al., 2018).

Prompts include (1) task instructions (see Appendix A), (2) 10 examples of input-output pairs illustrating desired behavior (see Appendix C), and (3) lemma-transcription pairs. For each language, examples were verified by native speakers. Responses include a series of morpheme-transcription pairs. Responses reporting a perplexity $>1.4$ were dropped. For each language, a random sample of 150 morphemes was drawn for verification by native speakers. Resulting error rates are shown in Table 1

[1]We do not consider shared submorphemic segments, such as ‘ct’ in ‘connection’ and ‘construction,’ to cause a transitivity confound because, while this study indirectly explores their potential semantic contributions, they currently lack any formally recognized independent semantic value.

[2]https://openai.com/index/gpt-4-1/

[3]https://openai.com/index/introducing-structured-outputs-in-the-api/

| Language | Errors | Error rate (95% CI) |
|---|---|---|
| English | 3/150 | 2.0% ± 2.24% |
| Spanish | 1/150 | 0.67% ± 1.3% |
| Hindi | 0/150 | 0% ± 0% |
| Finnish | 7/150 | 4.67% ± 3.38% |
| Turkish | 6/150 | 4.0% ± 3.14% |
| Tamil | 7/150 | 4.67% ± 3.38% |

Table 1: Segmentation error rates with 95% confidence intervals on random samples of 150 morphemes per language as verified by native speakers.

### 2.2.3 Embeddings

Semantic embeddings are retrieved with FastText (Bojanowski et al., 2016), which can embed morphemes as subword strings. Phonetic embeddings are retrieved by mean-pooling PanPhon feature vectors (Mortensen et al., 2016), dropping zero-variance dimensions in each language and dataset-normalizing values. (Mean-pooling involved taking the average component-wise value for a set of vectors.) Phonetic embeddings were verified in each language with similarity matrices of randomly sampled morpheme transcriptions. One sample of morphemes and their phonetic embeddings in English is shown in Figure 1.

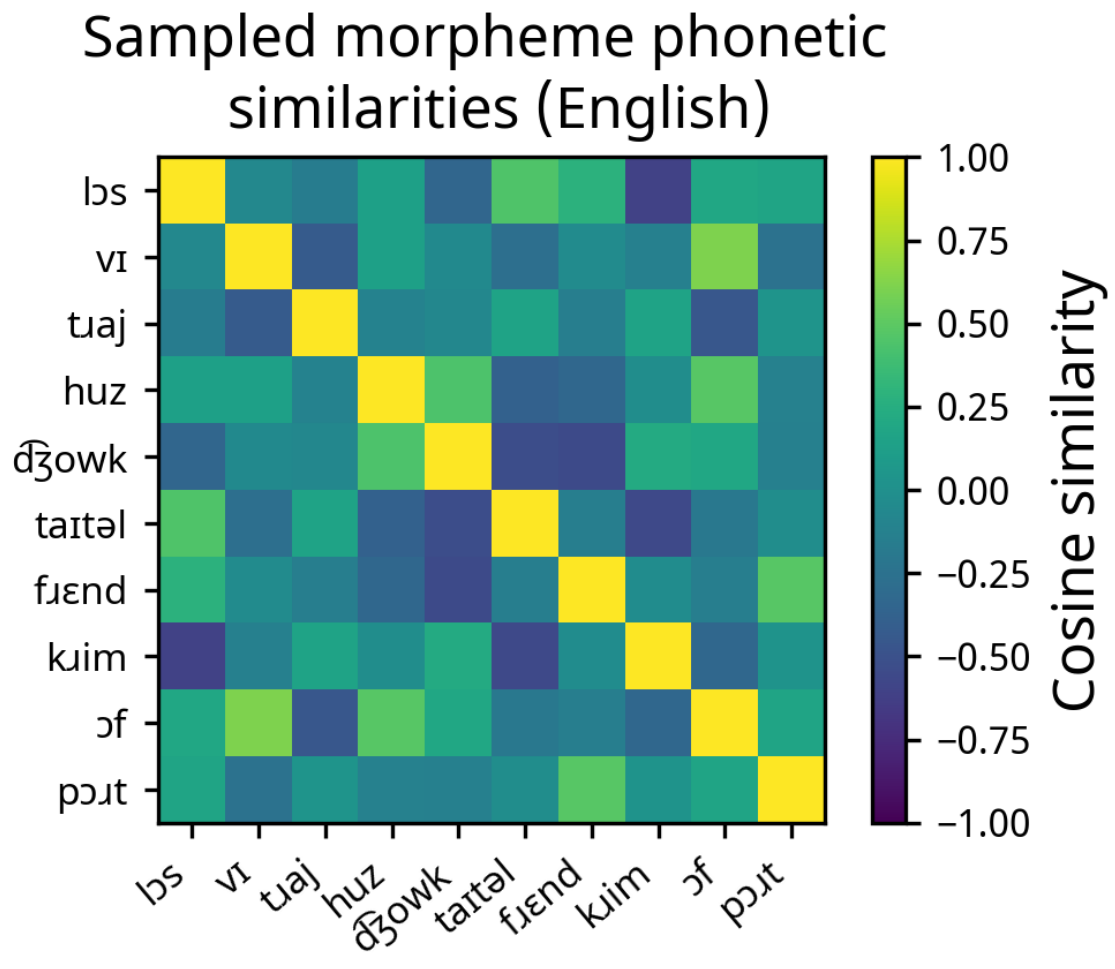

Figure 1: 10 randomly sampled English morpheme IPA transcriptions and their derived embeddings' phonetic similarities to each other.

## 2.3 Global analyses

For each language, we derive phonetic and semantic similarity matrices for its entire set of morphemes. We conduct several analyses on these two similarity matrices to investigate potential phonosemantic alignments at a global scale.

### 2.3.1 Representational Similarity Analysis

To test for global isomorphism between our phonetic and semantic similarity spaces, we conduct Representational Similarity Analysis (RSA; Kriegeskorte et al., 2008), which correlates two similarity matrices and quantifies the degree to which geometry of relationships is preserved across spaces. For each language, we compute Spearman's $\rho$ between phonetic and semantic similarity matrices aross all morpheme pairs. Significant positive correlations indicate global monotonic alignment between morphemes' phonetic and semantic similarities.

### 2.3.2 Mutual Information

Testing for monotonic relationships does not reveal potentially complex, nonlinear dependencies between phonetic and semantic spaces. To test for such relationships, we conduct Mutual Information (MI; Shannon, 1948) tests, which measures statistical dependence between two random variables without assuming linearity. For each language, we discretize similarities into 20 equal width bins and compute MI values between the discretized similarity distributions. Significant MI values indicate nonlinear statistical dependencies between morphemes' phonetic and semantic similarities.

### 2.3.3 $k$-Nearest-Neighbors overlap

Due to the size of our phonetic and semantic similarity spaces–and probable domination of arbitrary relationships at such a scale–we also evaluate potential dependencies at a local level. To do so, we evaluate overlap between each morpheme's $k$-Nearest-Neighbors ($k$NN overlap; Fix and Hodges, 1989; Easley and Kleinberg, 2010). For each morpheme in each language, we identify its 10 nearest neighbors in both phonetic and semantic spaces and calculate what proportion of neighbors are shared, averaging such proportions across all morphemes for a final scalar value. Significant $k$NN overlap values indicate the presence of shared phonosemantic neighborhoods across spaces.

### 2.3.4 Canonical Correlation Analysis

To identify phonosemantic alignments at a global scale made up of arbitrary combinations of features, we use Canonical Correlation Analysis (CCA; Hotelling, 1936), which finds linear combinations of sets of variables which drive alignment between the two spaces. Such maximally correlated linear combinations (canonical variates) be-tween phonetic and semantic features identify dimensions of maximal phonosemantic alignment, and can be manually inspected for interpretation. For each language, we extract the first 5 canonical variate pairs from phonetic and semantic embeddings, and compute Spearman's $\rho$ values for each pair. A significant correlation for a given canonical variate pair reveals an identified dimension of phonosemantic alignment.

#### 2.3.5 Canonical variate loadings and manual interpretation

For each such pair of canonical variates with a significant correlation, we examine loadings (the degree to which each original variable contributes to a given canonical variate) for all phonetic and semantic dimensions, and draw possible interpretations of each such pair of canonical variates. Because we use mean-pooling on character-wise feature vectors from PanPhon, phonetic dimensions are already interpretable, with each dimension corresponding to a specific phonological feature. To identify a phonetic pole, we collect feature dimensions from the top 75th percentile of positive or negative loading values depending on the pole, provided they reach a minimum threshold value of 0.05. Semantic dimensions, on the other hand, are not immediately interpretable, given that FastText uses dense embeddings. Instead of relying on these dimensions directly, we identify the directions of each pole of a given canonical variate in semantic space, and find the top 10 closest neighbors in FastText space which are also found in the WordFreq corpus above a Zipf score (Zipf, 1949) cutoff of 4.5. We then conduct manual inspections on the phonetic and semantic poles, and provide possible interpretations of the directions involved.

For all correlational analyses, we use Spearman's $\rho$ rather than Pearson's $r$ because a rank-based correlation would be resistant to outliers and distributional assumptions unfit to make about this linguistic data. Statistical significance is assessed via permutation testing with 1,000 shuffles per experiment. We use 500 points to create a null distribution. Repeated runs yielded stable results.

### 2.4 Subspace analyses

To investigate distributional realizations of some widely proposed phonosemantic scales in the literature, we conduct a separate set of analysis, where we define phonetic and semantic subspaces, project vocabularies onto them, and evaluate rank correlations. We investigate 5 such scales: magnitude-sonority, angularity-obstruency, fluidity-continuity, brightness-vowel frontness, and agility-phonological lightness.

To investigate alignment of phonetic and semantic subspace pairs, we adopt an approach inspired by Grand et al. (2022), where subspaces are defined with lines connecting the centroids of embeddings of two opposing sets of exemplars, and examine projections of items onto the subspace. In our case, we define two subspaces for each proposed phonosemantic scale: one in phonetic similarity space, the other in semantic similarity space. Subspaces are defined by the lines which connect the centroids of opposing exemplar set embeddings. Exemplars used to define phonetic subspaces are available in Appendix D Table 10; exemplars used to define semantic subspaces in each language are available in Appendix D Tables 11 and 12. Phonetic exemplars are PanPhon embeddings of single IPA phonemes and do not vary across languages.

For each subspace, we select the 10,000 words closest to the line by perpendicular Euclidian distance to it in FastText embeddings space. We note that we do not use our sets of morphemes for subspace analyses, because we are not calculating pairwise phonetic and semantic similarity scores, and thus do not have a transitivity confound to avoid. Phonetic embeddings are retrieved for these words using the same mean-pooling method described in 2.2.3. For each subspace pair, the selection of words is projected onto the phonetic and semantic subspaces, and rank correlations are calculated between projection coordinates using Spearman's $\rho$. Statistical significance is assessed via permutation testing with 1,000 shuffles per experiment with 5000 point null distributions. Poles are aligned such that a positive correlation supports a hypothesized alignment and vice versa.

## 3 Results

### 3.1 Global analyses

Numerical results of global analyses are available in Table 2. Across all 5 languages, none had significant RSA $\rho$ values, nor significant mutual information values. However, significant $k$NN overlap values were observed across all 5 languages, each with small numerical magnitude but high significance ($2\% - 3.9\%$ overlap, each with $p < 0.001$). Additionally across all languages, significant canonical rank correlations were identified in the first 5 canonical dimensions–except for English, which had an insignificant canonical rank correlation in the 5th canonical variate. Further, all languages but English demonstrate strong canonical rank correlations in the first canonical variate (all $> 0.5$ except for English at $0.376$). The first 3 canonical variates for all 5 languages demonstrate canonical rank correlation $p$-values below $0.001$, and most remain similarly high in the 4th and 5th canonical variates.

| Language | $n$ morphemes | RSA ($\rho$) | MI (bits) | $k$NN overlap | CCA CV1 ($\rho$) | CCA CV2 ($\rho$) | CCA CV3 ($\rho$) | CCA CV4 ($\rho$) | CCA CV5 ($\rho$) |
|---|---|---|---|---|---|---|---|---|---|
| English | 2153 | -0.027 | 0.001 | **0.020**$^{***}$ | **0.376**$^{***}$ | **0.318**$^{***}$ | **0.315**$^{***}$ | **0.176**$^{*}$ | 0.161 |
| Spanish | 1929 | 0.021 | 0.001 | **0.032**$^{***}$ | **0.598**$^{***}$ | **0.463**$^{***}$ | **0.299**$^{***}$ | **0.261**$^{***}$ | **0.224**$^{**}$ |
| Hindi | 1714 | -0.038 | 0.004 | **0.025**$^{***}$ | **0.554**$^{***}$ | **0.337**$^{***}$ | **0.303**$^{***}$ | **0.240**$^{**}$ | **0.197**$^{*}$ |
| Finnish | 1719 | 0.123 | 0.015 | **0.034**$^{***}$ | **0.519**$^{***}$ | **0.351**$^{***}$ | **0.304**$^{***}$ | **0.241**$^{**}$ | **0.213**$^{**}$ |
| Turkish | 1626 | 0.132 | 0.015 | **0.034**$^{***}$ | **0.538**$^{***}$ | **0.504**$^{***}$ | **0.305**$^{***}$ | **0.282**$^{***}$ | **0.229**$^{**}$ |
| Tamil | 1217 | 0.034 | 0.007 | **0.039**$^{***}$ | **0.538**$^{***}$ | **0.474**$^{***}$ | **0.408**$^{***}$ | **0.301**$^{***}$ | **0.269**$^{***}$ |

Table 2: Global analysis results. Significant values bolded, with levels: * $p < 0.05$, ** $p < 0.01$, *** $p < 0.001$.

| CV | Semantic Pole (+) | Phonetic Pole (+) | Semantic Pole (−) | Phonetic Pole (−) | Semantic Interpretation | Phonetic Interpretation |
|---|---|---|---|---|---|---|
| 1 | e, l, j, r, o, ma, y, f, h, le | Syllabic, Continuant, Sonorant, Voice, Strident | front, back, stuck, grab, stick, holding, pull, hanging, straight, down | Consonantal, Tense, Low, Labial, Lateral | Tensile/directional attachment | Tension |
| 2 | far, even, there, worse, though, only, that, fact, gone, one | Sonorant, Rounded, Coronal, Continuant, Voice | co, http, inc, id, e, ed, q, f, r, eu | Distributed, Syllabic, Delayed Release, Consonantal, Strident | Scalarity | Concentration |
| 3 | development, economic, importance, period, annual, region, aspects, final, historical, peak | Coronal, Anterior, Strident, Lateral, Continuant | ya, ok, if, told, me, lol, asked, know, i, oh | Back, Distributed, Syllabic, Voice, Labial | Informality | Ease of articulation |
| 4 | reported, agreed, stated, prior, approved, claimed, initial, identified, failed, previously | Anterior, Coronal, Consonantal, Strident, Lateral | kim, chris, alex, jesus, david, michael, daniel, kevin, asian, jim | Distributed, Sonorant, High, Delayed Release, Tense | Documentation | Constriction |

Table 3: English canonical variates' (CVs') semantic and phonetic loadings and provided interpretations. Semantic and phonetic interpreted values are proportional to one another. For example, in CV3, informality is proportional to ease of articulation.

Phonetic canonical variate loadings and words aligned with semantic canonical variate loadings are available in Table 3 for English and Appendix A for non-English languages. Phonetic and semantic poles derived via the method described in 2.3.5 are largely interpretable, though some pairs of poles only have one interpretable direction, with the other largely uninterpretable.

### 3.2 Subspace analyses

Numerical results of subspace analyses are available in Table 4. For 3 of the 5 proposed phonosemantic scales, a majority or one half of rank correlations are positive and significant, largely supporting the hypothesized existence of proposed phonosemantic scales. For example, 3 rank correlations were significant (each $p < 0.001$) and positive in the angularity–obstruency scale (of the famous *Kiki–Bouba* effect) and 3 were insiginificant. However, we also observe some results contradictory to these hypotheses: in 1 scale (fluidity–continuity), one half of the rank correlations are significant and negative, and only 1 is significant and positive; in the brightness–vowel scale, only 3 rank correlations are significant, though 2 of the 3 are positive. Overall, 12 rank correlations were significant and positive, 7 were significant and negative, and 11 were insignificant. 3 significant and positive correlations are shown in Figure 2.

| Language | Magnitude–Sonority | Angularity–Obstruency | Fluidity–Continuity | Brightness–Vowel frontness | Agility–Phonological lightness |
|---|---|---|---|---|---|
| English | **0.050**$^{***}$ | 0.009 | **0.021**$^{*}$ | -0.012 | 0.017 |
| Spanish | **-0.075**$^{***}$ | **0.111**$^{***}$ | **-0.088**$^{***}$ | **-0.025**$^{*}$ | **0.074**$^{***}$ |
| Hindi | **0.061**$^{***}$ | 0.008 | 0.000 | **0.028**$^{**}$ | **0.024**$^{*}$ |
| Finnish | 0.018 | **0.136**$^{***}$ | **0.105**$^{***}$ | **0.101**$^{***}$ | -0.001 |
| Turkish | **0.021**$^{*}$ | 0.011 | **-0.085**$^{***}$ | 0.002 | **-0.039**$^{***}$ |
| Tamil | 0.001 | **0.113**$^{***}$ | **-0.036**$^{**}$ | -0.006 | **-0.032**$^{**}$ |

Table 4: Spearman's $\rho$ correlations quantifying hypothesized alignments between word projections onto semantic scales and their putatively corresponding phonetic scales. Significant values bolded, with levels: * $p < 0.05$, ** $p < 0.01$, *** $p < 0.001$.

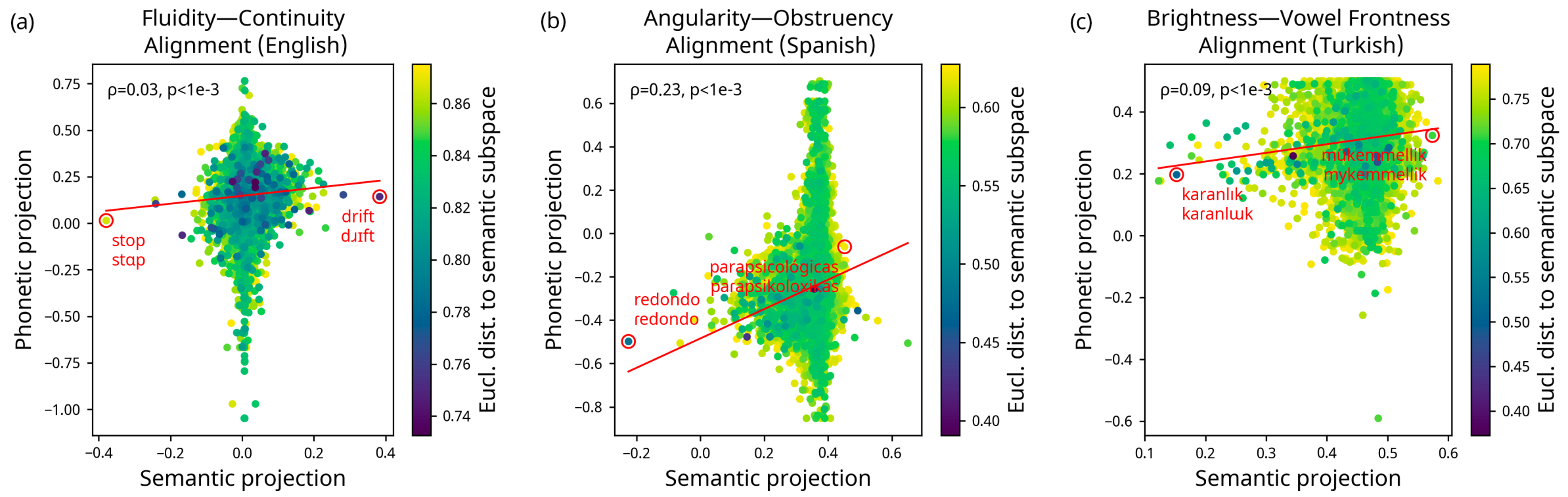

Figure 2: Examples of word projections onto paired semantic and phonetic subspaces. (a) English magnitude (fluidity) vs. sonority (continuity). (b) Spanish angularity vs. obstruency. (c) Turkish brightness vs. vowel frontness. Words tend to cluster more strongly in semantic projections than phonetic projections, likely as a result of a more extreme downprojection (from 300-dimensional semantic embeddings downprojected to 1-dimensional subspace compared to 16-21-dimensional phonetic embeddings downprojected to 1-dimensional subspace.)

## 4 Discussion

Overall, results indicate that phonosemantic iconicity operates primarily through specific dimensions and local neighborhoods rather than as a global, monotonic property across a given language. This indication upholds principles of arbitrariness writ large cooexisting with pockets of phonosemantic iconicity in language.

Negligible RSA and MI values in global analyses suggest an absence of broad isomorphism between the phonetic and semantic spaces over morpheme sets. Concurrently, significant $k$NN overlap values indicate the presence of local, neighborhood-level alignments across languages, and strong canonical rank correlation values indicate the presence dimensions of phonosemantic operation. Weaker results in English might be a product of diverse origins of derivational morphology (Blake, 2017; Smith, 2014).

### 4.1 Interpreted phonosemantic alignments across canonical variates

A variety of interesting phonosemantic alignments are indicated by canonical loadings analyses on observed significant canonical variates. In English, we identify 4 potential alignments: tensile/directional attachment–tension (CV1), scalarity–concentration (CV2), informality–ease of articulation (CV3), and documentation–constriction (CV4). Interestingly, some similarities in phonetic and semantic poles are observed across languages. For example, we identify an informality–ease of articulation scale in Finnish too (CV1). In some cases, identified dimensions have two interpretable poles. For example, the English informality–ease of articulation scale has an opposite pole associated with financial language semantically and ef-fortful articulation phonetically; in the Finnish rendition, the opposite pole is associated with white versus red political alignment[4] semantically, and effort of articulation phonetically.

Other such alignments have similarly interesting features: we identify a stillness–resonance scale in Hindi (CV3), which associates semantics of stillness and state of being with resonant phonetics. In the stillness pole, words such as ॐ ("om", a sacred mantric sound) and है (*is/are*) are juxtaposed with words such as स्टेशन (*station*), ट्रैक (*truck*), and फेंक (*throw*) at the opposite pole. Phonetic features such as [+continuant], [+strident], and [+voice] at the resonant pole are juxtaposed with [+consonantal], [+coronal], and [+delayed release] at the opposite pole. We encourage the reader to observe Table 3 and Appendix A for more interpretations, or to interpret the data directly themselves.

### 4.2 Subspace analyses

Rank correlations between projections onto phonetic and semantic subspaces generally support previous hypotheses in the literature on the existence of particular phonosemantic scales. Most notably, results for the angularity–obstruency scale indicate a strong positive alignment crosslinguistically. Interestingly, results for the fluidity–continuity scale show a negative alignment in most languages. Results for other scales indicate a degree of crosslinguistic variation, though still aligning with previous hypotheses generally.

### 4.3 Future work

With 6 languages, patterns of crosslinguistic variation in both global and subspace analyses are largely suggestive if not inconclusive; analysis of more languages would help clarify these patterns. Further, the analysis of more manually defined phonosemantic subspace alignments, evaluation of more canonical variates, and interpretation thereof across the 6 languages could reinforce drawn conclusions and suggested patterns.

Future work might also investigate other modalities of iconicity, such as graphosemantic or graphophonetic iconicity in logographic languages (see Wu et al., 2025; Sun et al., 2019) or sign language (see Perlman et al., 2018).

[4] https://en.wikipedia.org/wiki/Finnish_Civil_War

## 5 Conclusion

Our analyses across 6 typologically diverse languages consistently reveal pockets of phonosemantic iconicity in local neighborhoods or global dimensions, and support the presence of phonosemantic iconic effects in several previously proposed alignments across the literature. We discover an array of interpretable phonosemantic alignments in observed canonical variates across the 6 languages, and present both direct data and possible interpretations for each of them. We conclude that dimensions of phonosemantic iconicity do indeed permeate the languages involved in our investigation.

## Limitations

Global analyses could stand to improve from higher sample sizes, though we do not consider current sample sizes to be explicitly problematic for our task. Morpheme set sizes are limited by their growth rate with respect to number of words presented for segmentation (which decreases as the set is expanded) and by the cost of segmenting words with our few-shot learning-based approach, which also requires native-speaker verifications. These factors make scaling to new languages or larger sample sizes more difficult, and makes full replication less accessible than might be desired ideally.

Elsewhere in the pipeline, established linguistic tools are used, which are generally quite reliable, but can have problems with relatively low-resource languages, such as Tamil. No particular problems were observed, but differences in data quality are certainly plausible.

## Ethics statement

This work involves computational analysis of linguistic data and presents minimal ethical concerns. LLMs were employed as tools for morpheme segmentation, and they can energetically costly, but no direct societal harm is foreseen from this work.

# Appendix

## A Canonical variate interpretations for non-English languages

| CV | Semantic Pole (+) | Translation (+) | Phonetic Pole (+) | Semantic Pole (−) | Translation (−) | Phonetic Pole (−) | Semantic Interpretation | Phonetic Interpretation |
|---|---|---|---|---|---|---|---|---|
| 1 | art, new, the, francés, digital, barcelona, online, tipos, profesionales, adultos | art, new, the, French, digital, Barcelona, online, types, professionals, adults | Consonantal, Anterior, Labial, Coronal, Delayed Release | da, ve, ay, sí, mí, va, él, la, mi, sé | give, go, oh, yes, me, goes, he, the, my, I know | Syllabic, Tense, Sonorant, Continuant, Voice | commerciality | obstruction |
| 2 | xd, ah, eh, yo, ay, ok, es, x, oh, un | xd, ah, eh, me, oh, ok, it's, x, oh, a | Distributed, Back, Labial, Rounded, Voice | establecer, considerar, función, ofrecer, necesaria, existencia, asegurar, aspectos, permitir, presentar | establish, consider, function, offer, necessary, existence, ensure, aspects, allow, present | Coronal, Anterior, Strident, Nasal, Syllabic, Tense | management | pressure |
| 3 | tierra, cabeza, oriental, piedra, espalda, punta, interior, arena, montaña, fuego | earth, head, eastern, stone, back, tip, interior, sand, mountain, fire | Low, Coronal, Delayed Release, Sonorant, Voice | sé, ok, p, r, si, co, c, b, he, q | I know, ok, for, r, if, co, c, good, he, what | Distributed, High, Labial, Rounded, Syllabic, Tense | earthiness | resonance/ sonority |
| 4 | lenguaje, literatura, cultural, religión, cultura, industrial, carácter, ámbito, género, desarrollo | language, literature, cultural, religion, culture, industrial, character, scope, gender, development | High, Lateral, Continuant, Delayed Release, Sonorant | q, ahi, ah, me, yo, asi, he, despues, ok, ya | what, there, ah, me, I, like this, I have, then, ok, already | Nasal, Anterior, Labial, Distributed, Syllabic, Tense | literature/ culture | fluidity |
| 5 | f, b, v, iv, p, c, r, j, g, ii | f, b, v, iv, p, c, r, j, g, ii | Back, Syllabic, Tense, High, Rounded | quizás, porque, momento, siempre, final, mejor, peor, pasa, quizá, pienso | perhaps, because, moment, always, end, better, worse, passes, perhaps, I think | Labial, Nasal, Strident, Sonorant, Anterior | opinion | frontness |

Table 5: Spanish canonical variates' (CVs') semantic and phonetic loadings and provided interpretations. Semantic and phonetic interpreted values are proportional to one another. (For example, in CV3, *earthiness* is proportional to resonance/sonority. Where a given semantic pole is nonsensical, then (1) it is considered null and the other pole is used for interpretation and (2) translations are copies of those words. (For example, see CV2 or CV5 positive poles.)

| CV | Semantic Pole (+) | Translation (+) | Phonetic Pole (+) | Semantic Pole (−) | Translation (−) | Phonetic Pole (−) | Semantic Interpretation | Phonetic Interpretation |
|---|---|---|---|---|---|---|---|---|
| 1 | अंतर्राष्ट्रीय, अंतरराष्ट्रीय, निर्वाचन, भारतीय, विधानसभा, नगर, चिकित्सा, हिन्दी, राष्ट्रीय, शिक्षा | international, nationalization, Indian, parliament/legislative assembly, city, medical/treatment, national, education | Consonantal, Coronal, Low, Anterior, Strident | तो, वो, अब, तू | to, now, you | Syllabic, Back, Distributed, Spread Glottis, Voice | political/ national language | friction |
| 2 | र, स, क, म | ra, sa, ka, ma | Consonantal, Spread Glottis, Coronal, Strident | उसने, साहब, अरे, हमने, लेकिन, उन्हें, मैंने, यार, जिसने, मुझे | he/she (did), sir, hey/oh, we (did), but, to them, I (did), friend/-dude, as he/she (did), to me | Tense, Long, Syllabic, Sonorant, Voice | expression of frustration, informality | duration/ sonority |
| 3 | निकाल, बाहर, स्टेशन, ट्रैक, लाइन, काट, जगह, फेंक, डाल, आसानी | copy/ imitation, outside, station, truck, line, corner/edge, place, throw, lentils/branch, ease/easily | Consonantal, Lateral, Coronal, Spread Glottis, Delayed Release | है, अर्थात, ॐ, ऐ, हु, डॉ | is/are, place/location, om (sacred sound), hey/come, am | Syllabic, Continuant, Strident, Distributed, Voice | movement—stillness | channeling—resonance |
| 4 | द, एंड, अ, बी, वी, जे, एस, र | da, aing, a, bi, vi, je, es, ra | Strident, Tense, Syllabic, High | अंदर, भीतर, छोड़, बाहर, निकल, फिर, देश, मुझको, मुसलमान, वापस | inside, inside/within, leave/abandon, outside, exit/come out, again/then, country, to me, Muslim, back/return | Back, Low, Long, Consonantal, Anterior | migration/ nationalism | consonantality/ open back vowels |
| 5 | र, म, क, स, मा, नि | ra, ma, ka, sa, mother, no/name | Long, High, Low, Spread Glottis, Nasal | ऐसा, सारे, अपने, फिर, वही, दूसरे, हमेशा, वहाँ, दुनिया, कभी | like this/such, go/move, one's own, again/then, the same, others/second, always, there, world, ever/sometimes | Labial, Anterior, Continuant, Lateral, Consonantal | temporality | consonantal frontness |

Table 6: Hindi canonical variates' (CVs') semantic and phonetic loadings and provided interpretations. Semantic and phonetic interpreted values are proportional to one another. (For example, in CV5, *temporality* is proportional to consonantal frontness.) Interpreted values separated by an emdash (–) indicate poles of the scale. (For example, in CV3, *movement* is the positive pole of the interpreted semantic scale; *stillness* the negative pole.)

| CV | Semantic Pole (+) | Translation (+) | Phonetic Pole (+) | Semantic Pole (−) | Translation (−) | Phonetic Pole (−) | Semantic Interp. | Phonetic Interp. |
|---|---|---|---|---|---|---|---|---|
| 1 | jyväskylä, suomalainen, ilkka, david, kansainvälinen, punainen, poliittinen, amerikkalainen, valkoinen, sosiaalinen | Jyväskylä (city), Finnish, Ilkka (male first name), David, international, red, political, American, white, social | Delayed Release, Back, Labial, Distributed, Continuant | sä, ko, ni, ne, ai, mä, te, oi, ku, j | you (informal), ko (interrogative enclitic), so/yeah/then (shortened), they/those, ai (exclamation), I (informal), you all, oh (archaic), when/as/because (colloquial), j | Coronal, Nasal, Consonantal, Anterior, Strident | white versus red political alignment—informality | effort of articulation—ease of articulation |
| 2 | voi, ollut, tule, hyvä, tulee, ole, jää, tarvitse, pitäisi, turha | can/may, been, come, good, comes/will come, be, stay/remain/ice, need, should, pointless | Syllabic, Tense, Distributed, Continuant, High | in, to, or, and, the, as, a, by, for, my | in, to, or, and, the, as, a, by, for, my | Consonantal, Nasal, Coronal, Anterior, Strident | potentiality | high resonance |
| 3 | tuollainen, tämäkin, tällainen, sekin, mies, joku, sellainen, tottakai, kun, todellakin | that kind, this too, this kind, that too, man, someone, that kind (neutral distance), of course, when/as/because, indeed/really | Sonorant, Voice, Nasal, Syllabic, Tense | j, u, i, my, l, ko, il, to, y, la | j, u, i, my, l, ko (interrogative enclitic), il, to, y, la | Consonantal, Coronal, Rounded, Labial, Anterior | demonstrativity/ referentiality | sonority |
| 4 | ku, ko, ai, ni, j, o, mä, i, sä, oi | when/as/because (colloquial), ko (interrogative enclitic), ai (exclamation), so/yeah/then (shortened), j, is (colloquial), I (colloquial), i, you (colloquial), oh (archaic) | Rounded, High, Labial, Distributed, Back | lainkaan, tarpeeksi, suomalaista, tavallista, korkeintaan, kovaa, luonnollisesti, normaalisti, vähemmän, tarkoita | at all, enough, Finnish, usual, at most, hard/loud, of course, normally, less, mean/intend | Low, Delayed Release, Anterior, Coronal, Lateral | conversational tonality—stern tonality | softness—supralaryngeality |
| 5 | markku, poika, mies, yritys, vaimo, anne, isä, paavo, markus, opettaja | Markku (male first name), boy/son, man/husband, company/business/endeavor, wife, Anne (female first name), father, Paavo (male first name), Markus (male first name), teacher | High, Distributed, Delayed Release, Consonantal, Back | oo, en, ei, ok, no, se, be, in, is, n | is/are (colloquial), I don't, no/not/doesn't, OK, well/so (discourse particle), it/that, be, in, is, n | Strident, Continuant, Syllabic, Tense, Rounded | relational nominality—copulativity | affricativity—vocalic tension |

Table 7: Finnish canonical variates' (CVs') semantic and phonetic loadings and provided interpretations. Semantic and phonetic interpretative values are proportional to one another. (For example, in CV3, *potentiality* is proportional to high resonance.) Interpreted values separated by an emdash (–) indicate poles of the scale. (For example, in CV4, conversational tonality is the positive pole of the interpreted semantic scale; stern tonality the negative pole.)

| CV | Semantic Pole (+) | Translation (+) | Phonetic Pole (+) | Semantic Pole (−) | Translation (−) | Phonetic Pole (−) | Semantic Interp. | Phonetic Interp. |
|---|---|---|---|---|---|---|---|---|
| 1 | ba, da, la, sa, ı, a, un, ın, u, aç | ba, also/in/at, with, if, him/her/it, to, your, your, him/her/it, open | Back, Low, Rounded, Distributed, High | idare, organize, kabul, takdir, hizmet, analiz, talep, tarif, profesyonel, davet | administration/management, organize, admission/acceptance, recognition, service, analysis, request, recipe, professional, invitation | Anterior, Consonantal, Tense, Lateral, Nasal | institutionality/affiliation | consonantality |
| 2 | sigorta, normalde, amerikan, amerika, telefon, banka, araç, amerikalı, film, hastane | insurance, normally, American (adj), America, phone, bank, vehicle/-tool/means, American (noun), film, hospital | Low, Back, Consonantal, Anterior | ni, me, le, di, ba, ü, sa, te, i, ye | ni (accusative suffix), me (negation morpheme), with, di (past tense suffix), ba, ü, if, te (locative suffix), i (3rd person accusative suffix), ye (dative suffix) | High, Rounded, Voice, Distributed, Syllabic | sociocultural infrastructure | unclear |
| 3 | mustafa, bey, ismet, mehmet, ahmet, süleyman, kemal, ibrahim, muhammed, hakan | all proper nouns (male given names) | Sonorant, Nasal, Voice, Anterior, Labial | sıvı, plastik, saç, metal, kağıt, renk, ısı, deri, yağ, boya | liquid, plastic, hair/sheet metal, metal, paper, color, heat, leather/skin, oil/fat, paint/dye | Distributed, Strident, Delayed Release, Rounded, Consonantal | inanimism/materiality | obstruency |
| 4 | lar, yap, r, l, gir, ba, n, me, m, dır | lar (plural suffix), do/make, r (present tense marker), l (nominalizer), enter, look/begin, n (reflexive/passive suffix), me (negation suffix), m (1st person possessive suffix), dır (copular suffix) | Coronal, Anterior, Lateral, Voice, Delayed Release | güzel, gerçekten, şükür, güzeldi, hoş, bende, tatlı, öyle, fena, cidden | beautiful/nice, truly/really, gratitude/thanks, it was beautiful, pleasant, me too, sweet/cute, like that/so, bad, seriously/honestly | Distributed, Labial, Tense, Syllabic, Strident | affectivity | constriction |
| 5 | devamlı, fazla, daima, sürekli, daha, ayakta, oranda, içinde, ölçüde, dört | constantly/continuously, too much/excess, always, continuously, more/yet/still, upright, in proportion/at a rate, inside/within, in measure, four | Low, High, Back, Syllabic, Distributed | joe, mike, of, to, tv, vs, by, no, hd, 0 | joe, mike, of, to, tv, vs, by, no, hd, 0 | Anterior, Labial, Rounded, Nasal, Consonantal | temporal measurement | vowel backness |

Table 8: Turkish canonical variates' (CVs') semantic and phonetic loadings and provided interpretations. Semantic and phonetic interpretative values are proportional to one another.

| CV | Semantic Pole (+) | Translation (+) | Phonetic Pole (+) | Semantic Pole (−) | Translation (−) | Phonetic Pole (−) | Semantic Interp. | Phonetic Interp. |
|---|---|---|---|---|---|---|---|---|
| 1 | மெர்சல், பிக்பாஸ், சர்கார், ஆஃப், ஆப், பாஸ், பேக், சிங், மிஸ், டைம் | Astonished (Tamil film), Bigg Boss (reality TV show), Government (film), off, app, boss, back/return, sing, Miss (title), time | Consonantal, Strident, Long, Anterior, Labial | மன, து, நல, இட, ய, ம, நீ, தை, சம, அத | mind/mental, து (neuter suffix), well-being/health, place, ய (connector suffix), ம, you, Thai (Tamil month), equal/equilibrium, that | Sonorant, Voice, Syllabic, Continuant, Tense | ontologicality | sonority |
| 2 | கே, ஜி, ஜீ, பி, ஜே, of, அட, is, ஏ, வி | கே, ஜி, ஜீ, பி, ஜே, of, அட, is, ஏ, வி | Strident, Distributed, Delayed Release, Syllabic, Tense | கட்டி, போட்டு, களை, கொண்டு, குறி, குழாய், உற்பத்தி, கல், கரை, வந்து | bind/block/mass, put/install, weeds/remove/collections, with, mark/target, tube, production/manufacture, stone, shore/edge, came/coming | Rounded, High, Lateral, Labial, Consonantal | construction | constriction |
| 3 | யூ, ஓ, ஐ, ஜே, ஏ, டி, ஜி, x, டீ, it | யூ, ஓ, ஐ, ஜே, ஏ, டி, ஜி, x, டீ, it | Long, Tense, Syllabic, Continuant, Lateral | பக்தி, உணர்வு, ஆன்மீக, வேதனை, தெய்வ, பலன், சுகம், மகத்தான, வாழ்வு, பெரும் | devotion, emotion, spiritual, suffering, divine, result/fruit, comfort, great/grand, life, great/big | Anterior, Distributed, Labial, Consonantal, Delayed Release | spirituality | consonantality/affricativity |
| 4 | கள், அலி, செப், சர்மா, அல், தலைமையிலான, கான், சர்கார், ஆளுநர், சிங் | கள் (plural suffix), Ali (name), sep (abbr. of September?), Sharma (surname), Al (Arabic def. article), headed/led by, Khan (surname), Government (film), governor, Singh (surname) | Consonantal, Nasal, Low, Sonorant, High | நீ, அட, it, ஹி, ஏ, be, ஹா, டூ, 00, w | you (informal), அட (scolding exclamation), it, hi (informal), eh? (interjection), be, ஹா (laughter), டூ, 00, w | Syllabic, Tense, Rounded, Labial, Continuant | informality | roundness |
| 5 | ஜ.க, ச, ம, க, ந, ய, வ, ட, பொ, து | ஜ.க, ச, ம, க, ந, ய, வ, ட, பொ, து | Low, Strident, Delayed Release, Back, Continuant | தான், இன்று, அங்கு, ஆண்டு, இங்கு, அங்கே, நம், ஒரு, வாழும், இப்போது | self/itself, today, there, year, here, over there, our, one/a, who/that lives, now | Anterior, Nasal, Sonorant, Rounded, Long | referentiality (especially temporal) | sonority/length |

Table 9: Tamil canonical variates' (CVs') semantic and phonetic loadings and provided interpretations. Semantic and phonetic interpretative values are proportional to one another.

## B LLM system prompt for segmentation task

You are a meticulous linguistic expert tasked with breaking down a provided word in a given language into its absolute semantic primitives. This includes roots, bound morphemes, and any semantic primitives that contribute to the compositional meaning of the word. You will also be provided with the words phonetic transcription, and you are tasked with breaking up the phonetic transcription to align with the semantic decomposition. You will work in {lang}. Your fluency in {lang} is native and your linguistic knowledge PhD-level familiar. To reiterate: if **ever** a decomposition can be further decomposed, you have failed. These must not be further decomposable according to our rules. This is thus not about extracting morphemes, but rather pure semantic primitives. **You are not to return anything other than segments which can be found in the word or transcription–no modifications or functional descriptions of them. NOTHING other than the literal characters found in the word and its transcription should be returned.** Observe some examples below.
{examples}

## C Example sets for few-shot learning

### C.1 English examples

input: deconstruct,diːkənstrʌkt
(de,diː),(con,kən),(struct,strʌkt)

input: run,rʌn
(run,rʌn)

input: severance,sɛvərəns
(sever,sɛvər),(ance,əns)

input: unhappiness,ʌnhæpinəs
(un,ʌn),(happi,hæpi),(ness,nəs)

input: biodiversity,baɪoʊdaɪvɜrsəti
(bio,baɪoʊ),(divers,daɪvɜrs),(ity,əti)

input: microscopic,maɪkrəskɑpɪk
(micro,maɪkrə),(scop,skɑp),(ic,ɪk)
input: the,ðə
(the,ðə)

input: discontinuation,dɪskəntɪnjueɪʃən
(dis,dɪs),(con,kən),(tinu,tɪnju),(ation,eɪʃən)

input: hello,həloʊ
(hello,həloʊ)

input: shenanigans,ʃənænɪgənz
(shenanigan,ʃənænɪgən),(s,z)

### C.2 Spanish examples

input: desafortunadamente, desafoɾtunaða-mente
(des,des),(a,a),(fortuna,foɾtuna),(da,ða),(mente,mente)

input: zapatería,θapateɾia
(zapat,θapat),(ería,eɾia)

input: imprescindible,impɾesindiβle
(im,im),(pre,pɾe),(scind,sind),(ible,iβle)

input: sobremesa,soβɾemesa
(sobre,soβɾe),(mesa,mesa)

input: envejecer,embeχeθeɾ
(en,em),(vejec,beχeθ),(er,eɾ)

input: antepasados,antepasaðos
(ante,ante),(pas,pas),(ados,aðos)

input: contrarreloj,kontrarreloχ
(contra,kontra),(reloj,reloχ)

input: rascacielos,raskaθjelos
(rasca,raska),(cielos,θjelos)

input: el,el
(el,el)

input: perro,pero
(perr,per),(o,o)

input: gata,gata
(gat,gat),(a,a)"

### C.3 Hindi examples

input: किताब,kitaːb
(किताब,kitaːb)

input: लड़कियाँ,ləɽkijaːⁿ
(लड़की,ləɽkiː),(याँ,jaːⁿ)

input: जाऊँगा,d͡ʒaːuːⁿgaː
(जा,d͡ʒaː),(ऊँ,uːⁿ),(गा,gaː)

input: घरवाला,gʰərvaːlaː
(घर,gʰər),(वाला,vaːlaː)

input: अनुवादक,ənuvaːdək
(अनु,ənu),(वाद,vaːd),(क,ək)

input: विद्यालय,vidjaːləj
(विद्या,vidjaː),(लय,ləj)

input: सुनाई,sunaːiː
(सुन,sun),(आई,aːiː)

input: बेइज्जती,beɪd͡ʒd͡ʒətiː
(बे,be),(इज्जत,ɪd͡ʒd͡ʒət),(ई,iː)

input: खाकर,kʰaːkər
(खा,kʰaː),(कर,kər)

input: राजकुमारियों,raːd͡ʒkumaːrijoːⁿ
(राज,raːd͡ʒ),(कुमारी,kumaːriː),(यों,joːⁿ)

### C.4 Finnish examples

input: talo,tɑlo
(talo,tɑlo)

input: talossa,tɑlossɑ
(talo,tɑlo),(ssa,ssɑ)

input: kirjoista,kirjoistɑ
(kirja,kirjɑ),(i,i),(sta,stɑ)

input: lentokone,lentokone
(lento,lento),(kone,kone)

input: ymmärrän,ymmærræn
(ymmärrä,ymmærræ),(n,n)

input: opiskelisin,opiskelisin
(opiskel,opiskel),(isi,isi),(n,n)

input: työttömyys,tyøttømyys
(työ,tyø),(ttömyys,ttømyys)

input: juoksemassa,juoksemɑssɑ
(juokse,juokse),(ma,mɑ),(ssa,ssɑ)

input: kirjassani,kirjɑssɑni
(kirja,kirjɑ),(ssa,ssɑ),(ni,ni)

input: kuuntelemattomia,kuuntelemɑttomiɑ
(kuuntele,kuuntele),(ma,mɑ),(ttom,ttom),(ia,iɑ)

### C.5 Turkish examples

input: kitap,kitap
(kitap,kitap)

input: evler,evleɾ
(ev,ev),(ler,leɾ)

input: geliyorum,gelijoɾum
(gel,gel),(iyor,ijoɾ),(um,um)

input: başbakan,baʃbakan
(baş,baʃ),(bakan,bakan)

input: göremeyeceksiniz,gøɾemejed͡ʒeksiniz
(gör,gøɾ),(e,e),(me,me),(yecek,jed͡ʒek),
(siniz,siniz)

input: masadaki,masadaki
(masa,masa),(da,da),(ki,ki)

input: çiçekçi,t͡ʃit͡ʃekt͡ʃi
(çiçek,t͡ʃit͡ʃek),(çi,t͡ʃi)

input: vatandaşlık,vatandaʃlɯk
(vatan,vatan),(daş,daʃ),(lık,lɯk)

input: köprü,kœpɾy
(köprü,kœpɾy)

input: konuşamıyordum,konuʃamɯjoɾdum
(konuş,konuʃ),(a,a),(mı,mɯ),(yor,joɾ),(du,du),
(m,m)

### C.6 Tamil examples

input: வீடு,viːɖu
(வீடு,viːɖu)

input: புத்தகங்கள்,pʊt̪t̪əkəŋgəɭ
(புத்தகம்,pʊt̪t̪əkəm),(கள்,gəɭ)

input: மரத்தில்,məɾət̪t̪il
(மரம்,məɾəm),(இல்,il)

input: செல்கிறேன்,selgiɾeːn
(செல்,sel),(கிற்,giɾ),(ஏன்,eːn)

input: படிக்கவில்லை,pəɖikkəvilləi
(படி,pəɖi),(க்க,kkə),(இல்லை,villəi)

input: பார்த்துக்கொண்டிருந்தான், pɑːrt̪t̪ʊkkoɳɖirʊnd̪ɑːn
(பார்,pɑːr),(,t̪t̪ʊ),(கொண்டு,kkoɳɖ),
(இரு,irʊ),(ன்த்,nd̪),(ஆன்,ɑːn)

input: நிலச்சரிவு,niləʧʧəɾivʊ
(நிலம்,nilə),(சரிவு,ʧəɾivʊ)

input: நல்லவர்,nəlləvər
(நல்ல,nəllə),(அர,vər)

input: தமிழ்,t̪əmiɻ
(தமிழ்,t̪əmiɻ )

input: கற்றுக்கொடுத்தார்கள், kət̪t̪ʊkkoɖʊt̪t̪ɑːrgəɭ
(கல்,kə),(று,t̪t̪ʊ),(கொடு,kkoɖʊ),(த்,t̪t̪),
(ஆர்,ɑːr),(கள்,gəɭ)

# D Phonetic and semantic exemplars for subspace definitions in selected languages

| Scale | Positive Exemplars | Negative Exemplars |
|---|---|---|
| Magnitude-Sonority | ɑ, o, u, ɔ, ʊ | i, ɪ, e, ɛ |
| Angularity-Obstruency | p, t, k, tʃ | m, n, l, b, d, g |
| Fluidity-Continuity | l, m, n, r, f, v, s, z | p, t, k, b, d, g |
| Brightness-Vowel frontness | i, ɪ, e, ɛ | u, ʊ, o, ɔ |
| Agility-Phonological lightness | p, t, k, f, s, ʃ, i, ɪ | b, d, g, v, z, ʒ, a, ɑ |

Table 10: Phonetic exemplars used to define phonetic subspaces used in subspace analyses. Each subspace is defined as the line connecting the centroids of the positive and negative exemplar sets.

| Language | Magnitude-Sonority (+) | Magnitude-Sonority (-) | Angularity-Obstruency (+) | Angularity-Obstruency (-) |
|---|---|---|---|---|
| English | big, large, huge | small, tiny, little | sharp, pointed, angular | round, smooth, curved |
| Spanish | grande, enorme, gigante | pequeño, diminuto, chico | puntiagudo, afilado, angular | redondo, suave, curvo |
| Hindi | बड़ा, विशाल, विराट | छोटा, लघु, सूक्ष्म | नुकीला, तीखा | गोल, चिकना |
| Finnish | suuri, iso, valtava | pieni, pikkuinen, vähäinen | terävä, kulmikas, särmikäs | pyöreä, sileä, kaareva |
| Turkish | büyük, kocaman, iri | küçük, ufak, minik | sivri, keskin, köşeli | yuvarlak, pürüzsüz, kavisli |
| Tamil | பெரிய, மாபெரும் | சிறிய, குட்டி | கூர்மையான, முனையுள்ள | வட்ட, மென்மையான |

Table 11: Semantic exemplars for Magnitude-Sonority and Angularity-Obstruency subspaces. Each subspace is defined as the line connecting the centroids of the positive and negative exemplar sets.

| Language | Fluidity-Continuity (+) | Fluidity-Continuity (-) | Brightness-Vowel frontness (+) | Brightness-Vowel frontness (-) | Agility-Phonological lightness (+) | Agility-Phonological lightness (-) |
|---|---|---|---|---|---|---|
| English | flow, drift, glide | stop, jump, snap | bright, light, glow | dark, dim, shadow | fast, quick, swift | slow, heavy, lumbering |
| Spanish | fluir, flotar, deslizar | parar, saltar, romper | brillante, claro, luminoso | oscuro, tenue, sombra | rápido, veloz, ligero | lento, pesado, torpe |
| Hindi | बहना, तैरना | रुकना, कूदना | उजला, चमकदार | अंधेरा, मंद | तेज़, जल्दी, फुर्तीला | धीमा, भारी |
| Finnish | virrata, ajelehtia, liukua | pysähtyä, hypätä, napsahtaa | kirkas, vaalea, hohto | tumma, himmeä, varjo | nopea, pikainen, vikkelä | hidas, raskas, kömpelö |
| Turkish | akmak, sürüklenmek, kaymak | durmak, zıplamak, çatlamak | parlak, aydınlık, ışıltı | karanlık, loş, gölge | hızlı, çabuk, süratli | yavaş, ağır, hantal |
| Tamil | பாய், மிதந்து | நிறுத்து, தாவு | ஒளிர், பிரகாசமான | இருண்ட, மங்கலான | வேகமான, விரைவான | மெதுவான, கனமான |

Table 12: Semantic exemplars for Fluidity-Continuity, Brightness-Vowel frontness, and Agility-Phonological lightness subspaces. Each subspace is defined as the line connecting the centroids of the positive and negative exemplar sets.